\documentclass[11pt]{article}

\usepackage[margin=1in]{geometry}
\usepackage{authblk}
\usepackage{graphicx}
\usepackage[labelformat=parens,labelsep=space,font=small]{subcaption}
\usepackage{amsmath,amssymb}
\usepackage{booktabs}
\usepackage{array}
\usepackage{multirow}
\usepackage{xcolor}
\usepackage{enumitem}
\usepackage{placeins}
\usepackage[numbers,sort&compress]{natbib}

\usepackage{hyperref}
\hypersetup{
    colorlinks=true,
    linkcolor=blue,
    citecolor=blue,
    urlcolor=blue
}

\makeatletter
\newread\rc@probe
\newif\ifrc@stale
\def\rc@checkfile#1#2{
    \global\rc@stalefalse
    \openin\rc@probe=#1\relax
    \unless\ifeof\rc@probe
        \begingroup
        \endlinechar=-1
        \loop
            \readline\rc@probe to \rc@line
            \edef\rc@tmp{\noexpand\in@{\detokenize{#2}}{\rc@line}}\rc@tmp
            \ifin@ \global\rc@staletrue \fi
        \unless\ifeof\rc@probe
        \repeat
        \endgroup
    \fi
    \closein\rc@probe
}
\rc@checkfile{\jobname.aux}{abx@aux}
\ifrc@stale
    \AddToHook{begindocument/before}{%
        \let\rc@orig@input\@input
        \def\@input#1{\global\let\@input\rc@orig@input}%
    }
\fi
\rc@checkfile{\jobname.bbl}{biblatex}
\ifrc@stale
    \AtBeginDocument{%
        \let\rc@orig@bibliography\bibliography
        \def\bibliography#1{%
            \begingroup\let\@input@\@gobble\rc@orig@bibliography{#1}\endgroup}%
    }
\fi
\makeatother

\newcommand{\flowchart}[2][\textwidth]{%
    \IfFileExists{#2.pdf}{\includegraphics[width=#1]{#2.pdf}}{%
    \IfFileExists{#2.png}{\includegraphics[width=#1]{#2.png}}{%
    \IfFileExists{#2.jpg}{\includegraphics[width=#1]{#2.jpg}}{%
    \fbox{\parbox[c][4cm][c]{0.9\textwidth}{\centering
        Image not uploaded yet: \texttt{#2.png} (or \texttt{.pdf}, \texttt{.jpg})}}}}}%
}

\newcommand{\pkg}{\texttt{rc\_prediction}}

\newcommand{\corremail}{Ying-Cheng.Lai@asu.edu}

\newcommand{\blfootnote}[1]{%
    \begingroup
    \renewcommand{\thefootnote}{}%
    \footnotetext[0]{#1}%
    \endgroup
}

\title{%
    \vspace{-2.5em}%
    {RC Prediction: A Python Package for Parameter-Aware\\[0.15em]
    Reservoir Computing\par}%
}

\author[1]{Jin-Cheng Rao}
\author[1]{Zheng-Meng Zhai}
\author[1,2,\protect\hyperlink{corresponding-email}{$\ast$}]{Ying-Cheng Lai}

\affil[1]{%
    School of Electrical, Computer and Energy Engineering,
    Arizona State University, Tempe, AZ 85287, USA
}
\affil[2]{%
    Department of Physics,
    Arizona State University, Tempe, AZ 85287, USA
}

\date{\vspace{-2em}}

\begin{document}

\maketitle
\vspace{-1.5em}

\blfootnote{%
    $^{\ast}$ Corresponding author: {\color{blue}\hypertarget{corresponding-email}{\texttt{\corremail}}}
}

\section*{Summary}
A nonlinear dynamical system can be characterized by the autonomous
evolution of its state in time and its dependence on system parameters.
As a parameter varies, the dynamics can change quantitatively and, at bifurcation points, qualitatively. When the governing equations are unknown, a data-driven model should therefore capture both elements from observational data, rather than reproduce the dynamics at a single operating condition only. Parameter-aware reservoir computing (PARC) is a model-free framework for this purpose, where the
parameter is supplied to a reservoir computer through an additional input channel.  PARC is trained on time series recorded at several parameter values, and the
trained reservoir, run in closed loop as an autonomous dynamical system, generates
trajectories at parameter values that were never observed. \pkg{} is an open-source Python package that implements this workflow. It provides multi-parameter training of a reservoir, autonomous closed-loop prediction at user-specified parameter values, parameter scans, and a standard echo state network for forecasting at a fixed parameter. Hyperparameter optimization and optional GPU acceleration through PyTorch are also provided. We use three benchmark dynamical systems as examples to demonstrate its usage. As one of the featured applications, here we illustrate the prediction of critical transitions and system
collapse: trained only on pre-critical data, the model is queried beyond the training
range to predict whether the system collapses, to estimate the critical parameter $p_c^{*}$, and to characterize transient lifetimes.

\section*{Statement of need}

Most data-driven forecasting models of nonlinear dynamical systems, including conventional reservoir computers, are trained on time series recorded at a fixed parameter value and thus learn a single dynamical system. When the parameters change, the attractor of the underlying system can deform and, at a bifurcation, change its nature altogether. A model trained at one operating condition contains no representation of how the dynamics depend on the parameter; consequently, predicting the behavior at a new parameter value requires new data at
that value, which are often unavailable precisely in the regimes of greatest interest. What is needed is a
model that learns, from data at a limited number of parameter values, a family of
parameter-dependent dynamics that can be interrogated elsewhere in parameter space. Such a model enables, in principle, prediction of the dynamics at unseen parameter values by interpolation and extrapolation in parameter space; characterization of how dynamical behavior and its statistics vary with the parameter; analysis of qualitative
transitions and bifurcations; and forecasting of critical transitions and system
collapse. A model with these capabilities can serve as a digital twin of the dynamical system~\cite{zhai2026digital}.

Reservoir computing (RC)~\cite{jaeger2001,maass2002,lukosevicius2009} is a
recurrent neural network paradigm in which only a linear readout layer is trained, while
a large, randomly connected, fixed ``reservoir'' provides rich nonlinear dynamics~\cite{Lai:book}.
Because training reduces to a linear (ridge) regression problem, RC is computationally
cheap relative to fully trained recurrent networks and is well suited to data-driven
forecasting of chaotic systems, including spatiotemporally chaotic systems of large
effective dimension~\cite{pathak2018}. Moreover, a trained reservoir computer is itself a dynamical system: when its output is fed back as its next input, it evolves autonomously in closed loop and generates its own future trajectory without further measurements. In light of these features, reservoir computing has been widely utilized and studied in nonlinear dynamics, including prediction~\cite{jaeger2004,sussillo2009,pathak2017,pathak2018b,grigoryeva2018,griffith2019,carroll2019,haluszczynski2019,lymburn2019,jiang2019,vlachas2020,gonon2020,hart2020,chen2020,pandey2020,chattopadhyay2020,fan2020,klos2020,bollt2021,racca2021,platt2021,zhang2021,kim2021,gauthier2021,gilpin2023,patel2023,zhang2023,koglmayr2024,kong2024}, inference~\cite{lu2017,nakai2018,zimmermann2018,zhu2019,banerjee2019,chen2020b,lu2020,huang2020,gottwald2021,rohm2021,angelatos2021,fan2022,gauthier2022,li2024}, control~\cite{zhai2023control}, detection~\cite{carroll2018,zhai2023detect}, and physical implementations~\cite{fernando2003,appeltant2011,paquot2012,brunner2013,vandoorne2014,torrejon2017,du2017,fujii2017,larger2017,tanaka2019,rafayelyan2020,nakajima2020,zhong2022,milano2022,yan2024}. In addition, it has been combined with transformers~\cite{zhai2025transformer}, and machine learning of dynamical systems has been studied with meta-learning~\cite{zhai2025meta}. Generally, the ability of reservoir computing to learn the dynamics of a system has been deeply investigated, while how the dynamics change with the parameters can be modeled with reservoir computing remains an open question for standard RC.

Kong et al.~\cite{kong2021} extended standard RC to a \emph{parameter-aware} setting by feeding the parameter $p$ into the reservoir through an additional input channel and training a single readout on time series from several parameter values, named parameter-aware reservoir computing (PARC). The closed-loop
configuration supplies the autonomous evolution, and the parameter channel, which remains
open after training, turns the trained machine into a family of dynamical systems indexed
by $p$ that can be queried at any parameter value. Related studies have used this
framework to anticipate crises, system collapse, and transient lifetimes from pre-critical
data~\cite{kong2021}, to reproduce the statistics of transient chaos and intermittency
beyond the training regime~\cite{kong2021b}, to anticipate the onset of synchronization in
coupled oscillators~\cite{fan2021}, to construct digital twins that respond to driving
signals not encountered during training~\cite{kong2023}, to reconstruct bifurcation
diagrams of chaotic circuits from measured time series~\cite{luo2024}, and to track
time-varying parameters from partial state observation~\cite{zhai2024tracking}. Follow-up work has
also adapted the framework to update the reservoir readout online as new data
arrive~\cite{panahi2024}. PARC therefore addresses the problem of learning the dynamics together with their dependence on the parameter.

We note that several open-source libraries for reservoir computing already exist, such as ReservoirPy~\cite{trouvain2020} and PyRCN~\cite{steiner2022}, which provide flexible, general-purpose tools for building and training echo state networks. A parameter-aware model could in principle be assembled from such components, but these libraries focus on conventional time-series modeling and do not provide a dedicated workflow for training
a single model on time series pooled across several parameter values, closed-loop rollout at a prescribed parameter value, and the analyses built on such rollouts. We therefore developed \pkg{} to provide
this workflow as a documented, tested, and installable implementation, together with the benchmark systems and analysis tools needed to reproduce and extend the results of
Kong et al.~\cite{kong2021}.

While PARC has been applied to a wide range of dynamical tasks, here we choose one of its featured applications for demonstration, that is, the prediction of \emph{critical transitions}. Specifically, as a control parameter $p$ is varied smoothly past a threshold $p_c$, the long-term
behavior of the system changes abruptly, often from sustained and bounded dynamics to
transient chaos followed by escape or collapse. Examples include predator-prey collapse in ecological networks, voltage collapse in power grids, and regime shifts in climate subsystems~\cite{scheffer2009}. Throughout, $p_c$ denotes this true critical point, a
property of the system itself, which is generally unknown in practice. Its estimate
obtained from the trained reservoir alone is denoted by $p_c^{*}$. For benchmark systems,
where $p_c$ is known, the difference between $p_c^{*}$ and $p_c$ measures the accuracy
of the prediction. For this task, \pkg{} provides dedicated tools: collapse criteria that distinguish physical collapse from numerical divergence of
the surrogate, critical-point scans that report their finite horizon and the bracket of the transition, and ensembles over reservoirs with right-censored transient lifetimes.

\section*{Method}

\begin{figure}[htbp]
\centering
\flowchart{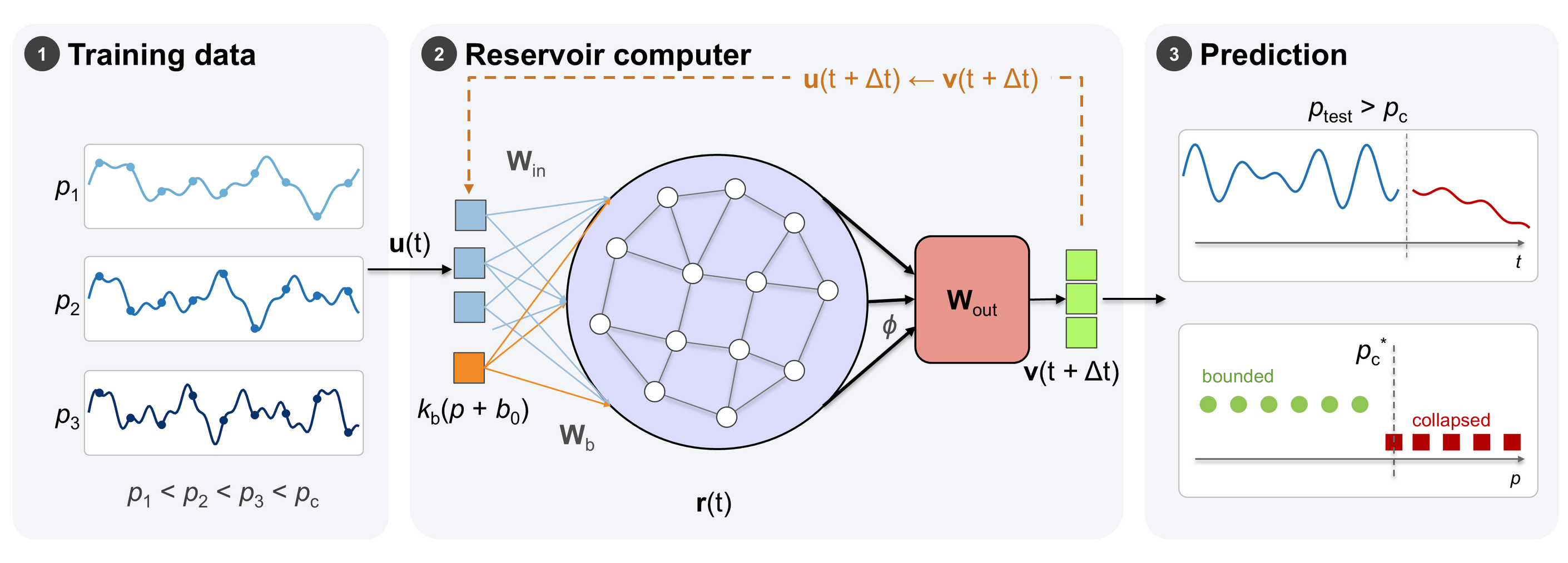}
\caption{Parameter-aware reservoir computing as implemented in \pkg{}, shown for the
collapse-prediction setting. (1) Time series are recorded at several pre-critical parameter
values $p_1 < p_2 < p_3 < p_c$. (2) The state $\mathbf{u}(t)$ (blue) enters the reservoir
through $\mathbf{W}_{in}$, and the scaled parameter $k_b(p + b_0)$ (orange) through its own
input weights $\mathbf{W}_b$. The reservoir, with state
$\mathbf{r}(t)$, is fixed and randomly connected; only the linear readout
$\mathbf{W}_{out}$ is trained, by a single ridge regression pooled over all training values
$p_j$. For prediction, the output is fed back as the next input,
$\mathbf{u}(t+\Delta t) \leftarrow \mathbf{v}(t+\Delta t)$ (dashed loop), while the
parameter input is held at a test value $p_{\mathrm{test}}$. (3) Top: the closed-loop
trajectory at one $p_{\mathrm{test}} > p_c$ is the model's prediction of whether and when
the system collapses. Bottom:
repeating this over a grid of parameter values gives the estimate $p_c^{*}$, the first collapsing grid value.}
\label{fig:pipeline}
\end{figure}

The PARC method proceeds in two stages. In the training stage, a single reservoir computer
is driven by time series recorded at several parameter values, each presented together
with its parameter value, and one readout is fit to all of them. The shared reservoir and
readout thus represent a family of parameter-dependent dynamical responses; the model is a
dynamical surrogate rather than a classifier of outcomes. In the prediction stage, the
trained reservoir is closed into an autonomous dynamical system, and the parameter channel
is used to query it at a chosen parameter value. Analyses such as collapse detection and
critical-point scanning are subsequently applied to the trajectories generated in this
way. Figure~\ref{fig:pipeline} illustrates the workflow in the collapse-prediction setting
featured here: the reservoir is trained at several pre-critical parameter values and then
run autonomously at unseen parameter values to detect collapse and estimate the critical
point. The bottom plot in the prediction panel of Fig.~\ref{fig:pipeline} is a binary scan that records, at each
grid value, only whether the closed-loop trajectory collapses within the horizon. It
locates a parameter-dependent transition but, unlike a bifurcation diagram, does not
resolve the attractor at each parameter value.

\subsection*{Parameter-aware reservoir dynamics}

Let $\mathbf{u}(t) \in \mathbb{R}^d$ denote the observed state of the dynamical system
and let $p$ denote the scalar bifurcation (control) parameter. Following
Kong et al.~\cite{kong2021}, the parameter enters the reservoir through its own input
channel, and the reservoir state $\mathbf{r}(t) \in \mathbb{R}^{N}$ evolves under a leaky
echo-state update

\begin{equation}
\mathbf{r}(t+\Delta t) = (1-\alpha)\,\mathbf{r}(t) + \alpha \tanh\big[\,
\mathbf{W} \cdot \mathbf{r}(t) 
+ \mathbf{W}_{in} \cdot \mathbf{u}(t) + k_b \mathbf{W}_b\,(p + b_0)\big],
\label{eq:reservoir-update}
\end{equation}

where $\mathbf{W} \in \mathbb{R}^{N \times N}$ is a fixed, sparse, symmetric, randomly
generated recurrent weight matrix (nonzero entries uniform in $[-1, 1]$, mean degree
$\langle k \rangle$) rescaled to a target spectral radius $\rho(\mathbf{W})$;
$\mathbf{W}_{in} \in \mathbb{R}^{N \times d}$ and $\mathbf{W}_b \in \mathbb{R}^{N}$ are
fixed dense input weights with entries uniform in $[-\sigma, \sigma]$, where $\sigma$ is
the input-scaling factor; the gain $k_b$ and bias $b_0$ scale and shift the parameter so
that its input is comparable in magnitude to that of the state; and $\alpha \in (0,1]$ is
the leaking rate. $\mathbf{W}$, $\mathbf{W}_{in}$, and $\mathbf{W}_b$ are drawn at random
once and never trained.

The readout uses a quadratic feature map that squares every other reservoir unit, a
trick introduced by Lu et al.~\cite{lu2017} and adopted by Kong et al.~\cite{kong2021}
that breaks the odd symmetry of $\tanh$ and improves the readout's ability to
reconstruct even observables:

\begin{equation}
\phi(\mathbf{r})_i =
\begin{cases}
r_i^2 & i \text{ odd (0-based indexing)} \\
r_i & \text{otherwise,}
\end{cases}
\qquad
\mathbf{v}(t+\Delta t) = \mathbf{W}_{out}\, \phi(\mathbf{r}(t+\Delta t)).
\label{eq:readout}
\end{equation}

Because $\mathbf{r}(t+\Delta t)$ has received the inputs up to and including
$\mathbf{u}(t)$, the readout $\mathbf{v}(t+\Delta t)$ is a one-step-ahead prediction. The
output weight matrix $\mathbf{W}_{out}$ is the only trained component of the model. It is
fit by ridge (Tikhonov-regularized) linear regression with regularization strength
$\lambda$ so that, during teacher-forced training, $\mathbf{v}(t+\Delta t) \approx
\mathbf{u}(t+\Delta t)$.

\subsection*{Training on multiple parameter values}

The key idea that allows the model to generalize across $p$ is that training data are drawn
from \emph{several distinct parameter values} rather than from a single trajectory. In the
general PARC framework, these values sample the parameter region of interest, and the
trained model may be queried between them (interpolation) or beyond them (extrapolation).
In the collapse-prediction setting considered here, all training values lie in the
pre-critical regime, $p_1 < p_2 < \dots < p_m < p_c$, and the model is queried beyond the
largest of them. For each training parameter $p_j$, a segment of the
corresponding time series is teacher-forced through Eq.~\eqref{eq:reservoir-update}
(i.e., the true $\mathbf{u}(t)$ is fed back at every step rather than the model's own
prediction), and the resulting $\big(\phi(\mathbf{r}(t+\Delta t)),\,\mathbf{u}(t+\Delta t)\big)$
pairs from all $m$ parameter values are pooled into a single ridge regression that fits
one shared $\mathbf{W}_{out}$. The first $w$ points of each
segment (the washout; $w = 10$ by default) are excluded from the regression targets, so that
transients caused by the reservoir's zero initial state do not enter the fit. When a
training length $L$ is set, the $L + 1$ points that follow are fitted. Optionally, Gaussian noise
of standard deviation $\sigma_n$ is added to the reservoir inputs (but not to the targets)
during training; this standard regularization makes the autonomous closed-loop dynamics
markedly more stable. The package requires at least three distinct training values
($m \geq 3$), all of which must be pre-critical, and at least 20 time steps per value.
This requirement reflects the collapse-prediction protocol for which the current
implementation is designed; it is a property of \pkg{}, not of the PARC methodology in
general, which does not restrict the training values to one side of a transition.

\subsection*{Closed-loop prediction and collapse detection}

Once trained, the model is queried at a test parameter $p_{\mathrm{test}}$ -- which may
lie between, below, or above the training values and, in the collapse setting, above
$p_c$ -- by running the reservoir in \emph{closed loop}: the model's own
one-step prediction is fed back as the next input state,
\begin{equation}
\mathbf{u}(t+\Delta t) \leftarrow \mathbf{v}(t+\Delta t),
\end{equation}
with $p_{\mathrm{test}}$ held fixed in the parameter input of Eq.~\eqref{eq:reservoir-update} at every step. This
autonomous rollout is the model's forecast of how the true system would behave at
$p_{\mathrm{test}}$, despite never having observed data at that parameter value. Whether
this forecast is faithful is not guaranteed and must be assessed, for instance against
ensembles of reservoirs or, in benchmarks, against the true system. For a chaotic system,
a rollout can be expected to follow the true trajectory only for a limited time; beyond
that, the relevant comparison concerns statistical and qualitative properties, such as
whether and when the trajectory leaves the attractor.

Collapse detection is a downstream analysis of these rollouts. Each rollout is run for a
finite horizon of $H$ steps, and collapse is detected according to a system-specific
criterion; the earliest event and its reason are reported:
\begin{itemize}[nosep]
    \item \textbf{Bound:} any state component exceeds a user-specified magnitude (used for
    the Ikeda map, whose post-critical trajectories escape to a distant region).
    \item \textbf{Extinction:} the density of a designated species falls below a threshold
    (0.01 by default), used for population and food-web systems.
    \item \textbf{Divergence:} the surrogate produces non-finite values or its state norm
    grows beyond ten times its initial scale. This is reported separately because a
    runaway surrogate is not, by itself, evidence of physical collapse.
\end{itemize}
Because no post-critical data enter the training, the predicted collapse is to be read as
an event: the state that the surrogate reaches after collapse lies outside the support of
the training data and is not expected to reproduce the physical post-collapse
state~\cite{kong2021}.

The trained model supports three kinds of prediction. A single closed-loop rollout at one
test parameter reports the trajectory and whether, when, and why collapse occurred. A scan
runs one rollout per value on a grid of test parameters. Because a finite rollout only
establishes whether collapse was \emph{observed within $H$ steps}, and transients become
long near $p_c$, the scan reports its horizon and the bracket between the last bounded and
the first collapsing grid value, flags non-monotonic outcomes, and uses the first
collapsing value as the estimate $p_c^{*}$ of $p_c$. Finally, an ensemble repeats the prediction
over independently initialized reservoirs and reports the fraction that collapse and the
distribution of transient lifetimes. Runs that do not collapse within the horizon are
treated as right-censored rather than assigned a lifetime, and the mean lifetime is taken
over collapsing runs only. Variation across reservoirs measures model uncertainty; it is
distinct from variation across initial conditions of the physical system, which the user
controls through the warm-up trajectory. By default, a rollout, including each rollout of a
scan, is initialized with the last 50 steps of the training series at the largest training
value not exceeding $p_{\mathrm{test}}$, or at the smallest training value if
$p_{\mathrm{test}}$ lies below all of them.

\section*{Package Structure}

\begin{figure}[htbp]
\centering
\flowchart{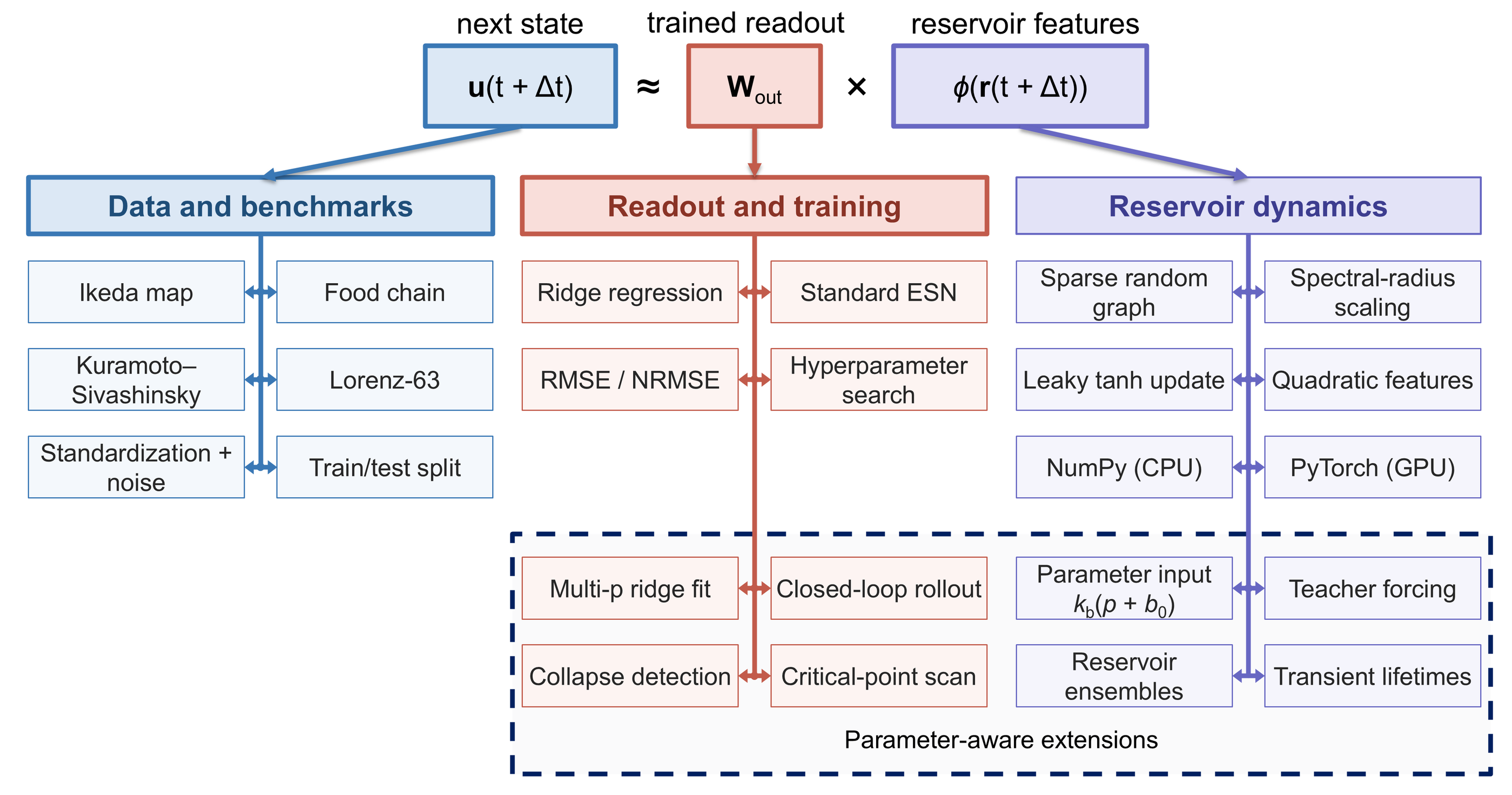}
\caption{Structure of \pkg{}. The model predicts the next state as a trained linear
readout of the reservoir features, $\mathbf{u}(t+\Delta t) \approx \mathbf{W}_{out}\,
\phi(\mathbf{r}(t+\Delta t))$; each factor of this equation corresponds to one part of the package,
which supplies the data $\mathbf{u}$, fits the readout $\mathbf{W}_{out}$, or generates the
reservoir features $\phi(\mathbf{r})$. The dashed box labeled ``Parameter-aware extensions'' collects the
components added for the parameter-aware framework on top of a standard echo state network.}
\label{fig:structure}
\end{figure}

\pkg{} is organized around the two factors of the readout equation,
$\mathbf{u}(t+\Delta t) \approx \mathbf{W}_{out}\,\phi(\mathbf{r}(t+\Delta t))$, as shown in
Fig.~\ref{fig:structure}: data sources and benchmark systems supply $\mathbf{u}$, the
reservoir components generate the features $\phi(\mathbf{r})$, and the readout components
fit and evaluate $\mathbf{W}_{out}$. The parameter-aware model reuses the same reservoir
and ridge-readout machinery as the standard ESN and adds the parameter channel,
multi-parameter training, and closed-loop rollout at a prescribed parameter value; collapse
detection, critical-point scans, and ensembles are analysis routines built on top of these
rollouts. The package is built with Hatchling, follows a standard source layout, and
depends only on NumPy, SciPy, and scikit-learn.

\subsection*{Optional GPU acceleration}

The recurrent update in Eq.~\eqref{eq:reservoir-update} is applied sequentially in time,
and its per-step cost is dominated by the dense matrix--vector product
$\mathbf{W} \times \mathbf{r}(t)$ (during readout training, the $\mathbf{X}^{\top} \times \mathbf{X}$
Gram matrix formed from stacked reservoir states), both of which scale with the square of
the reservoir size $N$. Both the standard ESN and the parameter-aware model accept an
opt-in GPU flag that offloads these operations to a CUDA device through PyTorch instead of
NumPy and SciPy on the CPU; both paths use double precision. Requesting the GPU without
PyTorch or a CUDA-visible device raises an error immediately rather than silently falling
back to the CPU, and with the flag off PyTorch is never imported.

Table~\ref{tab:gpu-bench} reports the wall-clock time of 500 reservoir state updates driven
by a fixed input sequence, measured with a benchmark script distributed with the package
on an Intel Core i9-13900KS (24 cores, 32 threads; NumPy with OpenBLAS using 24 threads)
and an NVIDIA GeForce RTX 4090 (CUDA 12.4, PyTorch 2.6.0). The recurrent matrix
$\mathbf{W}$ (10\% density) is stored as a SciPy sparse matrix on the CPU and as a dense
matrix on the GPU; in both cases the work per step scales with $N^2$. Each run time is
the median of five repetitions after a warm-up. Weight generation, including the
spectral-radius rescaling, is excluded and timed once; GPU timings include the host--device transfers that the package
performs at every step, each of which synchronizes the device. Because the recurrence is
sequential, the GPU only pays off once the per-step product is large enough to amortize
this overhead: in this configuration the GPU was slower for $N \leq 1000$ and faster for
$N \geq 2000$, by a factor of 4.5 at $N = 8000$. These figures depend on the hardware and
should not be read as general thresholds.

\begin{table}[htbp]
\centering
\caption{CPU vs.\ GPU wall-clock time for 500 reservoir state updates driven by a fixed
input sequence, at reservoir size $N$. Run times are medians of five repetitions. Setup
(a single measurement, excluded from the run times) is the one-time weight generation on
the CPU and is dominated at large $N$ by the spectral-radius computation. Hardware and
software are given in the text.}
\label{tab:gpu-bench}
\begin{tabular}{@{}rrrrr@{}}
\toprule
$N$ & Setup (s) & CPU run (s) & GPU run (s) & Speedup \\
\midrule
200  & 0.016 & 0.004 & 0.055 & $0.07\times$ \\
500  & 0.084 & 0.008 & 0.051 & $0.16\times$ \\
1000 & 0.118 & 0.020 & 0.052 & $0.38\times$ \\
2000 & 0.392 & 0.075 & 0.053 & $1.41\times$ \\
4000 & 1.791 & 0.287 & 0.116 & $2.47\times$ \\
8000 & 33.4  & 1.484 & 0.330 & $4.50\times$ \\
\bottomrule
\end{tabular}
\end{table}

With identical seeds, CPU and GPU reservoir states driven by the same input agree to
within $6 \times 10^{-16}$ over 500 steps ($N = 1000$). The fitted readouts agree to about
$10^{-7}$, because the two matrix-inversion routines differ. Autonomous closed-loop rollouts
of chaotic systems amplify such differences, as any change in floating-point summation
order would, and CPU and GPU trajectories separate beyond $10^{-6}$ after about a dozen
steps; only statistical properties, such as collapse probabilities, should be compared
between devices.

\section*{Installation and Usage}

\begin{figure}[htbp]
\centering
\flowchart{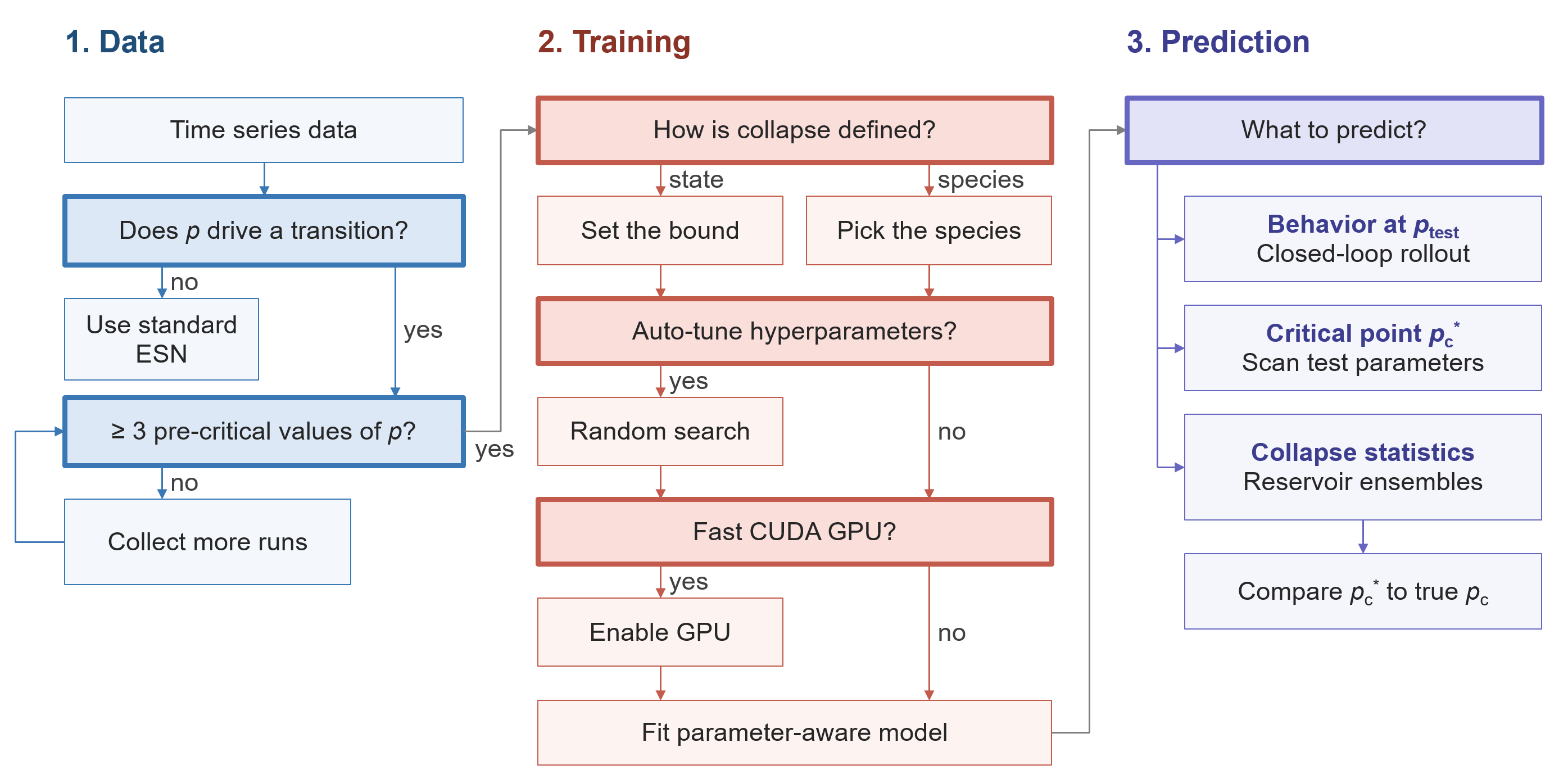}
\caption{Decision flow chart for applying \pkg{}. (1) Data: starting from a dataset, a
user decides between the standard ESN and the parameter-aware model and checks that enough
pre-critical data are available. (2) Training: the user configures collapse detection,
hyperparameter tuning, and GPU acceleration, and then fits the model. (3) Prediction: the
user selects the prediction that answers the scientific question.}
\label{fig:decision}
\end{figure}

\pkg{} requires Python $\geq 3.10$ and is published on the Python Package Index (PyPI)
and on GitHub under the Apache License 2.0. All results in this paper were produced with
version 0.1.3, the first release under this license; earlier releases were distributed
under the MIT License. Optional extras install Matplotlib and Jupyter for the examples and
tutorial notebook, development tools for testing and packaging, and PyTorch for GPU
acceleration. The package follows scikit-learn conventions: a model is constructed with
its hyperparameters, fit to training data, and then queried for predictions.
Figure~\ref{fig:decision} summarizes how a user can start from a dataset and choose the
appropriate model, collapse criterion, and prediction method, and
Table~\ref{tab:hyperparams} lists the hyperparameters of the parameter-aware model with
their default values.

\begin{table}[htbp]
\centering
\caption{Hyperparameters of the parameter-aware model and their default values.}
\label{tab:hyperparams}
\begin{tabular}{@{}lll@{}}
\toprule
Symbol & Meaning & Default \\
\midrule
$N$ & Reservoir size & 500 \\
$\langle k \rangle$ & Mean degree of the sparse recurrent matrix $\mathbf{W}$ & 4.0 \\
$\rho(\mathbf{W})$ & Spectral radius of $\mathbf{W}$ & 0.9 \\
$\sigma$ & Input scaling of $\mathbf{W}_{in}$ and $\mathbf{W}_b$ & 1.0 \\
$k_b$ & Parameter-channel gain & 0.5 \\
$b_0$ & Parameter-channel bias & 0.0 \\
$\alpha$ & Leaking rate & 0.3 \\
$\lambda$ & Ridge regularization strength & $10^{-8}$ \\
$\sigma_n$ & Training input-noise standard deviation & 0 \\
$w$ & Washout (initial points of each series excluded from the fit) & 10 \\
\bottomrule
\end{tabular}
\end{table}

Hyperparameters can also be selected automatically by a random search over user-specified
ranges, scored by prediction error on held-out data. Every prediction returns a small
result object: a single rollout reports the predicted trajectory, the test parameter, and
whether and at which step collapse occurred; a scan reports the parameter grid, the
collapse outcome at each grid point, and the estimate $p_c^{*}$; and an ensemble reports
the fraction of realizations that collapsed, their individual transient lifetimes, and
their mean.

\section*{Benchmark Systems}

\pkg{} bundles three benchmark dynamical systems, following the original paper and related
literature. Each system can simulate a single trajectory or a complete training set at
several parameter values. The Ikeda map and the food chain both change, at a critical
parameter value, from sustained bounded dynamics to transient chaos followed by escape or
collapse, and they are the systems on which the parameter-aware workflow is evaluated in
the worked examples. The Kuramoto--Sivashinsky equation is provided as a simulator of
spatiotemporally chaotic training data; no parameter-aware prediction results for it are
reported in this paper.

\subsection*{Ikeda map}

The Ikeda map originates as a discrete-time model of the transmitted light field in a
ring-cavity optical resonator~\cite{ikeda1979} and is a discrete-time complex map with
bifurcation parameter $\mu$:

\begin{equation}
z_{n+1} = \mu + \gamma z_n \exp\!\left(i\left(\kappa - \frac{\eta}{1+|z_n|^2}\right)\right),
\end{equation}

with default parameters $\gamma = 0.9$, $\kappa = 0.4$, $\eta = 6.0$, and critical value
$\mu_c = 1.0027$. The state is represented by the real and imaginary parts of $z_n$, and
the default training values $\mu \in \{0.91, 0.94, 0.97\}$ are all pre-critical. Below
$\mu_c$ the chaotic attractor satisfies $|\mathrm{Re}\,z|, |\mathrm{Im}\,z| < 2.5$; above it,
trajectories are transiently chaotic and eventually escape to a distant state with
$|z| \approx 5$.

\subsection*{Food chain}

A three-species Hastings--Powell / McCann--Yodzis food-chain model
\cite{hastings1991,mccann1995} with resource $R$, consumer $C$, and predator $P$
populations,
\begin{equation}
\begin{aligned}
\dot R &= R\left(1 - \frac{R}{K}\right) - \frac{x_c y_c\, C R}{R + R_0}, \\
\dot C &= x_c C\left(\frac{y_c R}{R + R_0} - 1\right) - \frac{x_p y_p\, P C}{C + C_0}, \\
\dot P &= x_p P\left(\frac{y_p C}{C + C_0} - 1\right),
\end{aligned}
\end{equation}
with $x_c = 0.4$, $y_c = 2.009$, $x_p = 0.08$, $y_p = 2.876$, $R_0 = 0.16129$, and
$C_0 = 0.5$, integrated with SciPy's adaptive Runge--Kutta solver and sampled at
$\Delta t = 1$. The carrying capacity $K$ is the bifurcation parameter, with critical
value $K_c \approx 0.99976$ and default training values $K \in \{0.97, 0.98, 0.99\}$.
Post-critical simulations first settle onto the pre-critical attractor during a warm-up
phase before the parameter is changed, following the transient-chaos control
methodology of Dhamala and Lai~\cite{dhamala1999}.

\subsection*{Kuramoto--Sivashinsky equation}

A one-dimensional Kuramoto--Sivashinsky (KS) equation~\cite{kuramoto1976,sivashinsky1977}
in the form used in the supplementary material of Kong et al.~\cite{kong2021},
\begin{equation}
\partial_t u = -\alpha\left(\partial_x^2 u + u\,\partial_x u\right) - \nu\,\partial_x^4 u,
\end{equation}
with $\nu = 4$ on a periodic domain of length $\pi$ discretized by 32 grid points. It is
solved with a fourth-order exponential time-differencing Runge--Kutta (ETDRK4) spectral
scheme with step $5 \times 10^{-6}$, and $\alpha$ is the bifurcation parameter with default
training values $\alpha \in \{196, 197, 198\}$. The KS system provides a spatiotemporally
chaotic benchmark beyond the low-dimensional Ikeda map and food chain, and is a standard
testbed for reservoir-computing forecasting of high-dimensional chaos~\cite{pathak2018}.

\begin{figure}[htbp]
\centering
\includegraphics[width=0.8\textwidth]{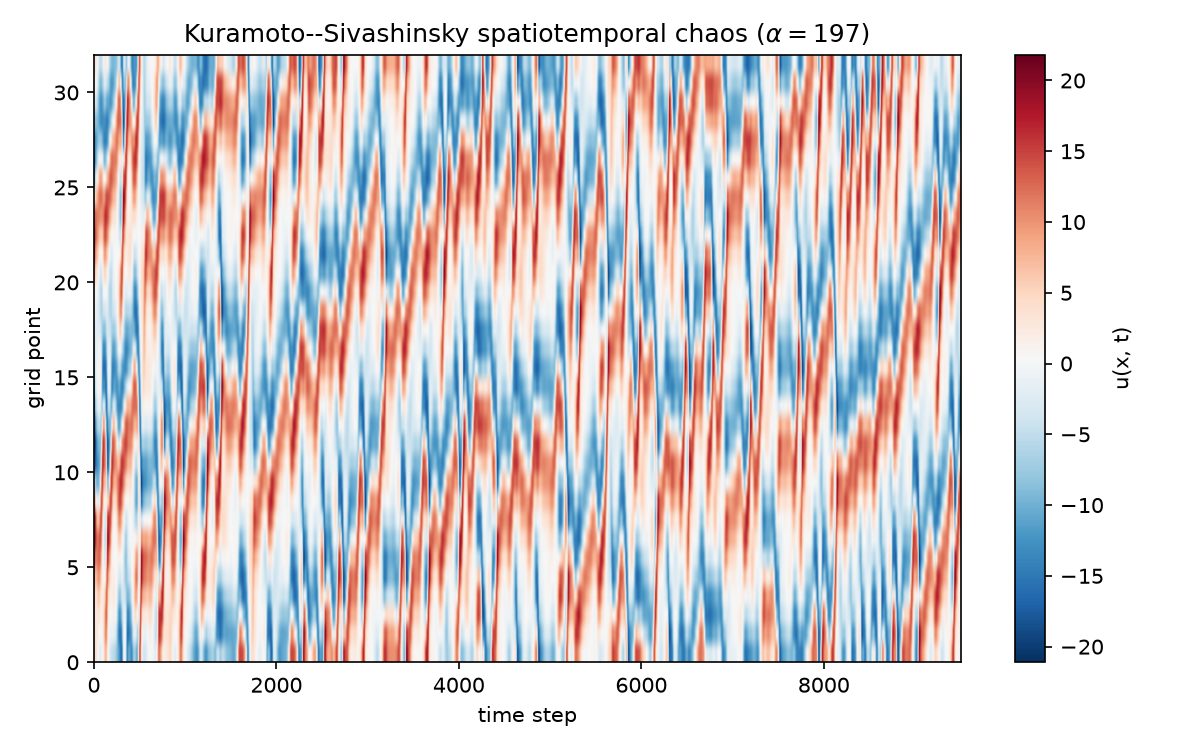}
\caption{Spatiotemporal chaos generated by the Kuramoto--Sivashinsky benchmark at
$\alpha = 197$, shown as a space--time heatmap of $u(x,t)$ over the 32-point periodic
grid.}
\label{fig:ks-spacetime}
\end{figure}

\section*{Worked Examples}

Example scripts in the repository reproduce every result below with version 0.1.3 (see
``Reproducibility'' below); Table~\ref{tab:example-settings} lists every setting, including
seeds, horizons, and initialization. The two parameter-aware examples, the Ikeda map and the
food chain, address a particularly demanding task: anticipating a transition that lies
beyond the training parameter range, from pre-critical data alone. They therefore test
extrapolation in parameter space, and they evaluate the collapse outcome within a finite
horizon, the estimated critical point, and transient lifetimes; they do not test the
reconstruction of bifurcation diagrams. The Lorenz-63 example uses the standard ESN at a
fixed parameter value and serves only as a check of the core reservoir implementation.

\begin{table}[htbp]
\centering
\caption{Settings used in the worked examples. Symbols follow
Table~\ref{tab:hyperparams}. The Ikeda hyperparameters are those selected by the blind
search (rounded here; exact values are stored with the scripts). For the parameter-aware
models, the first $w$ points of each series are excluded from the fit and the next $L + 1$
points are fitted.}
\label{tab:example-settings}
\small
\begin{tabular}{@{}llll@{}}
\toprule
 & Ikeda map (blind) & Food chain & Lorenz-63 \\
\midrule
Model & Parameter-aware RC & Parameter-aware RC & Standard ESN \\
Training data & $\mu \in \{0.91, 0.94, 0.97\}$ & $K \in \{0.97, 0.98, 0.99\}$ & $\Delta t = 0.02$ \\
Training length $L$ & 5000 per value & 400 per value & 3999 steps \\
Reservoir size $N$ & 500 & 400 & 500 \\
Connectivity & $\langle k \rangle = 6.29$ & $\langle k \rangle = 4$ & density 0.1 \\
Spectral radius $\rho(\mathbf{W})$ & 0.401 & 2.3 & 0.9 \\
Input scaling $\sigma$ & 2.17 & 3.6 & 0.5 \\
Parameter channel $k_b$, $b_0$ & 1.81, 0.175 & 0.50, $-2.2$ & --- \\
Leaking rate $\alpha$ & 0.770 & 0.30 & 0.3 \\
Ridge $\lambda$ & $9.0 \times 10^{-4}$ & $3 \times 10^{-5}$ & $10^{-8}$ \\
Training noise $\sigma_n$ & 0.0127 & 0 & --- \\
Washout $w$ & 10 points & 10 points & 500 steps \\
Collapse criterion & $\max_i |u_i| > 2.89$ & $P < 0.01$ & --- \\
Horizon $H$ & 5000 (up to 20000) & 1500; scan 1200 & --- \\
Scan grid & $[0.98, 1.03]$, step 0.002 & $[0.995, 1.005]$, 11 points & --- \\
Reservoir seeds & 1--24 & 3 & 42 \\
Initialization & last 50 training steps & last 50 training steps & zero state \\
Test data & scan grid & $K = 0.99$, $1.01$; scan grid & held-out last 20\% \\
\bottomrule
\end{tabular}
\end{table}

\begin{table}[htbp]
\centering
\caption{Blind hyperparameter search for the Ikeda example: sampling ranges (600
candidates, search seed 2026) and scoring protocol. Only pre-critical training data are
used.}
\label{tab:tuning}
\small
\begin{tabular}{@{}ll@{}}
\toprule
Hyperparameter & Range \\
\midrule
Reservoir size $N$ & $\{200, 300, 500\}$ \\
Mean degree $\langle k \rangle$ & uniform $[2, 8]$ \\
Spectral radius $\rho(\mathbf{W})$ & uniform $[0.1, 1.5]$ \\
Input scaling $\sigma$ & uniform $[0.3, 3]$ \\
Parameter gain $k_b$ & uniform $[0.2, 3]$ \\
Parameter bias $b_0$ & uniform $[-1.2, 0.5]$ \\
Leaking rate $\alpha$ & uniform $[0.3, 1]$ \\
Ridge $\lambda$ & log-uniform $[10^{-8}, 10^{-2}]$ \\
Training noise $\sigma_n$ & log-uniform $[10^{-3.5}, 10^{-1.3}]$ \\
\midrule
Data split & first 4000 steps of each training series for fitting, remainder held out \\
Score & Wasserstein-1 distance between $|z|$ in a 5000-step rollout and in held-out data; \\
 & 1 if the rollout crosses the bound; mean over 3 training values and 4 seeds \\
Collapse bound & $1.5 \times$ largest state component in the training data ($= 2.89$) \\
\bottomrule
\end{tabular}
\end{table}

\begin{figure}[htbp]
\centering
\begin{subfigure}[t]{0.49\textwidth}
\centering
\includegraphics[width=\linewidth]{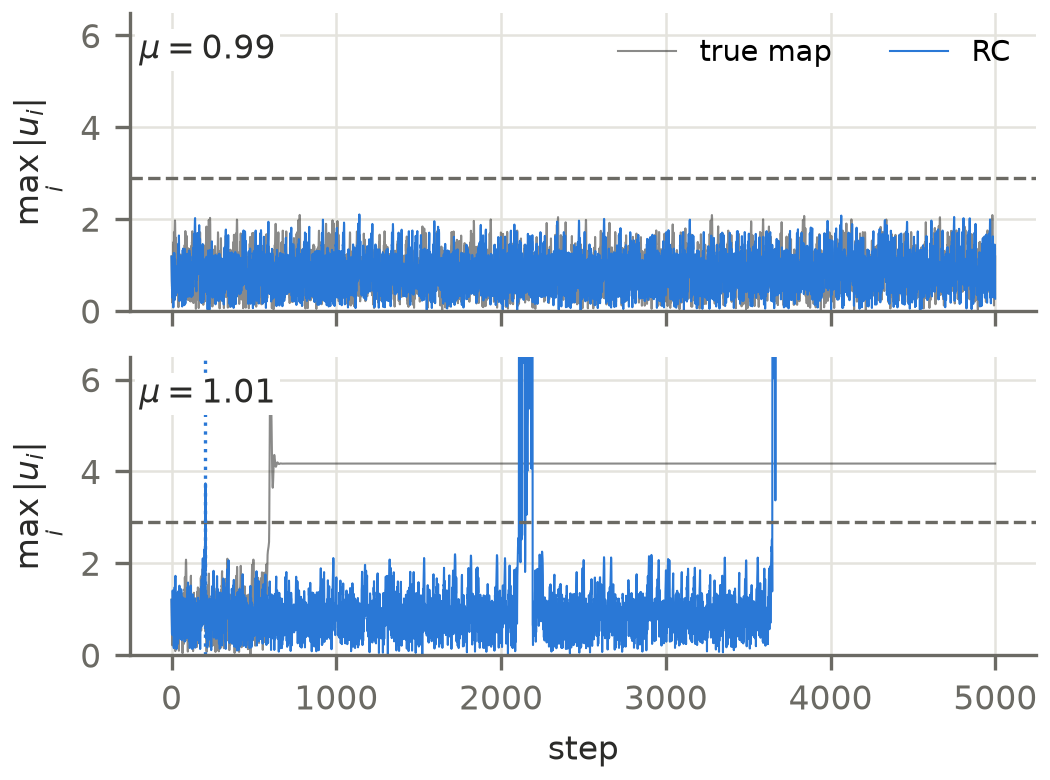}
\caption{Closed-loop prediction}
\end{subfigure}\hfill
\begin{subfigure}[t]{0.49\textwidth}
\centering
\includegraphics[width=\linewidth]{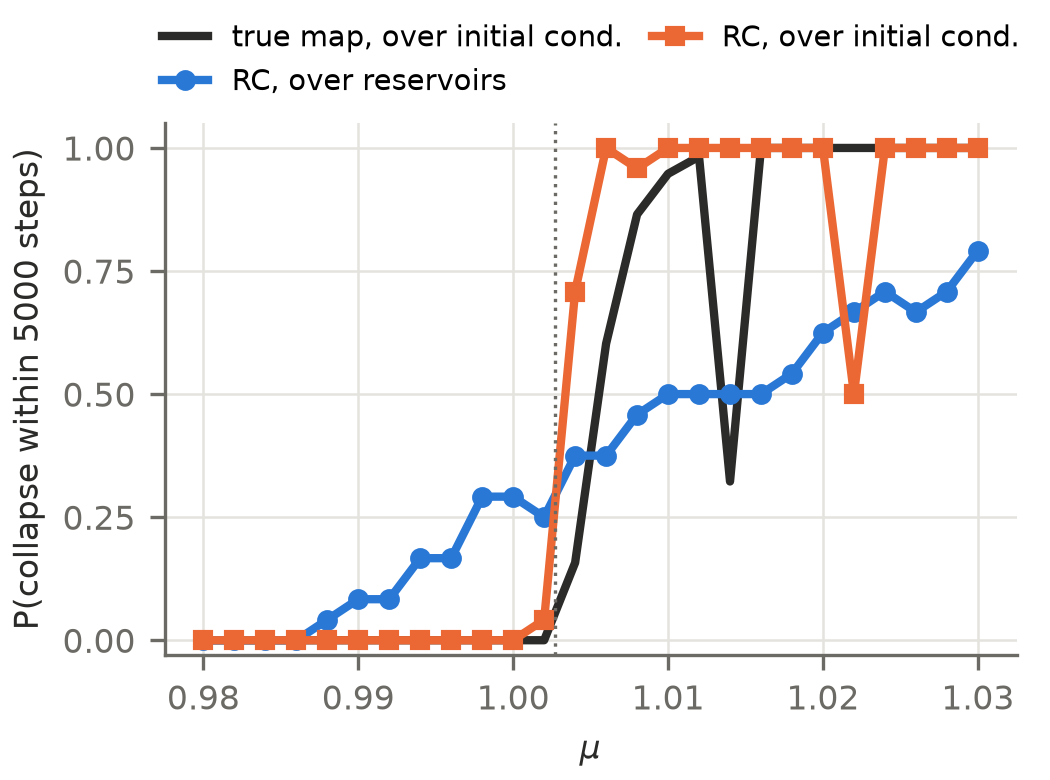}
\caption{Collapse probability}
\end{subfigure}

\medskip
\begin{subfigure}[t]{0.49\textwidth}
\centering
\includegraphics[width=\linewidth]{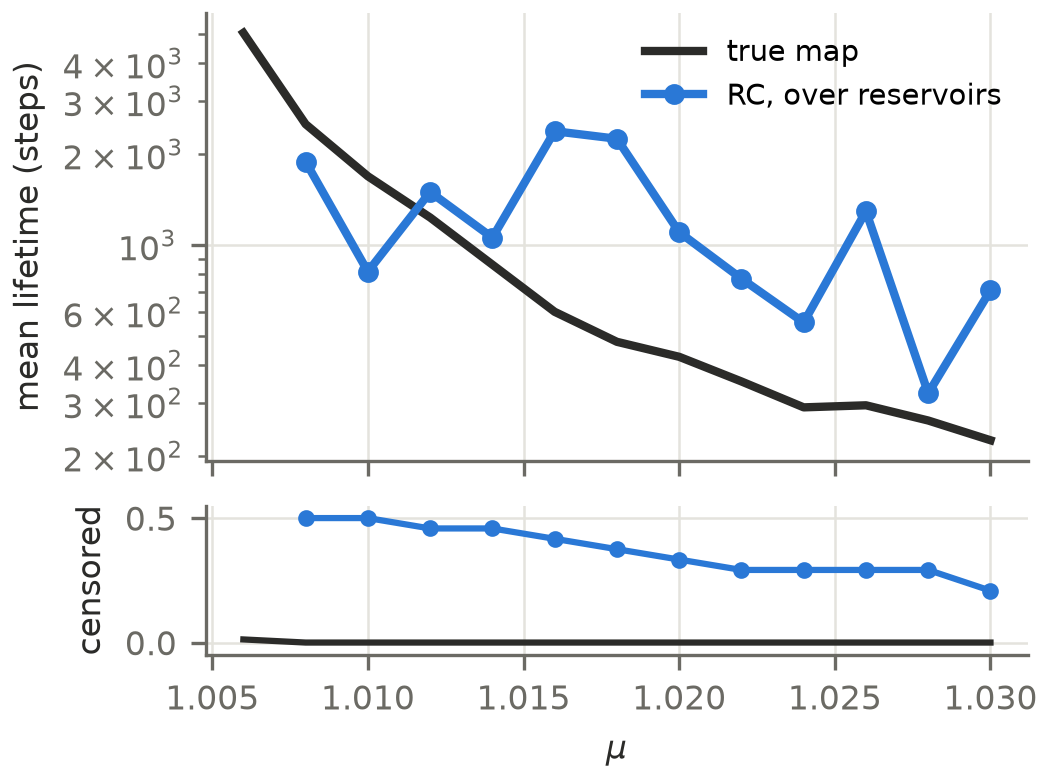}
\caption{Runs collapsing within 20000 steps}
\end{subfigure}\hfill
\begin{subfigure}[t]{0.49\textwidth}
\centering
\includegraphics[width=\linewidth]{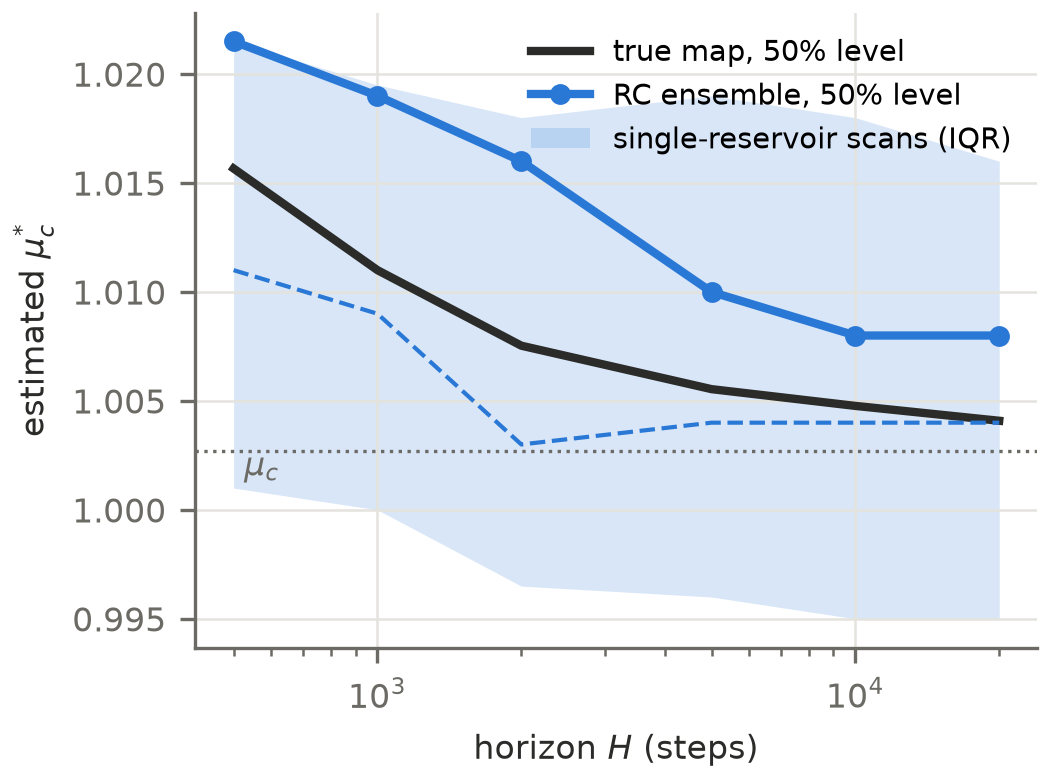}
\caption{Critical-point estimate vs.\ horizon}
\end{subfigure}
\caption{Prediction for the Ikeda map; collapse means $\max_i |u_i| > 2.89$. (a) Closed-loop prediction of reservoir seed 4 (blue), whose estimate is closest
to the median of 24 reservoirs, and the true map (gray) from the same initial state; the
dotted line marks the first threshold crossing. (b) Collapse probability within $H = 5000$
steps for the true map over 400 initial conditions (black), RC over 24 reservoirs (blue),
and the reservoir of (a) over 24 initial conditions (orange). The true curve dips at a
periodic window near $\mu = 1.014$; the orange curve dips similarly near $1.022$. (c) Mean
lifetime of runs collapsing within $H = 20000$ steps, where at least half collapse (top),
and the censored fraction (bottom). (d) Critical-point estimates versus $H$: 50\% collapse
level of the true map (black) and the RC ensemble (blue, solid), and median (blue, dashed)
and interquartile range (shading) of single-reservoir $\mu_c^{*}$. Dotted lines in (b) and
(d) mark $\mu_c$.}
\label{fig:ikeda-validation}
\end{figure}

\paragraph{Ikeda map:} All choices for this example are made from
pre-critical data alone: the collapse bound is 1.5 times the largest state component in the
training data ($|u_i| > 2.89$), and the hyperparameters come from a random search over 600
configurations (Table~\ref{tab:tuning}). Figure~\ref{fig:ikeda-validation} compares the
resulting model with the true map. The reservoir in panel (a) stays on
the attractor at $\mu = 0.99$ and collapses at $\mu = 1.01$. Across the ensemble, however,
the collapse probability rises gradually: up to 29\% of the reservoirs already collapse just
below $\mu_c$, where the true map never does, and 50\% is reached only at
$\mu \approx 1.010$ instead of $1.006$. Single-reservoir
scans are noisy (median 1.004, interquartile range $[0.996, 1.019]$; 5 of 24 reservoirs
never collapse on the grid), and the finite horizon biases even the true 50\% point upward.
Transient lifetimes are right only in order of magnitude: 820 and 710 steps at
$\mu = 1.01$ and $1.03$, against 1690 and 230 for the true map, with 21--50\% of the runs
not collapsing within the horizon. The surrogate also often crosses the bound briefly
before escaping, unlike the true map. None of the 578 detected collapses was a numerical
divergence.

With hyperparameters tuned instead against the true map's collapse outcomes, information
unavailable in practice, the ensemble 50\% point moves to $\mu \approx 1.001$.
Hyperparameter selection is therefore a dominant source of uncertainty in the predicted
critical point.

\begin{figure}[htbp]
\centering
\begin{subfigure}[t]{0.49\textwidth}
\centering
\includegraphics[width=\linewidth]{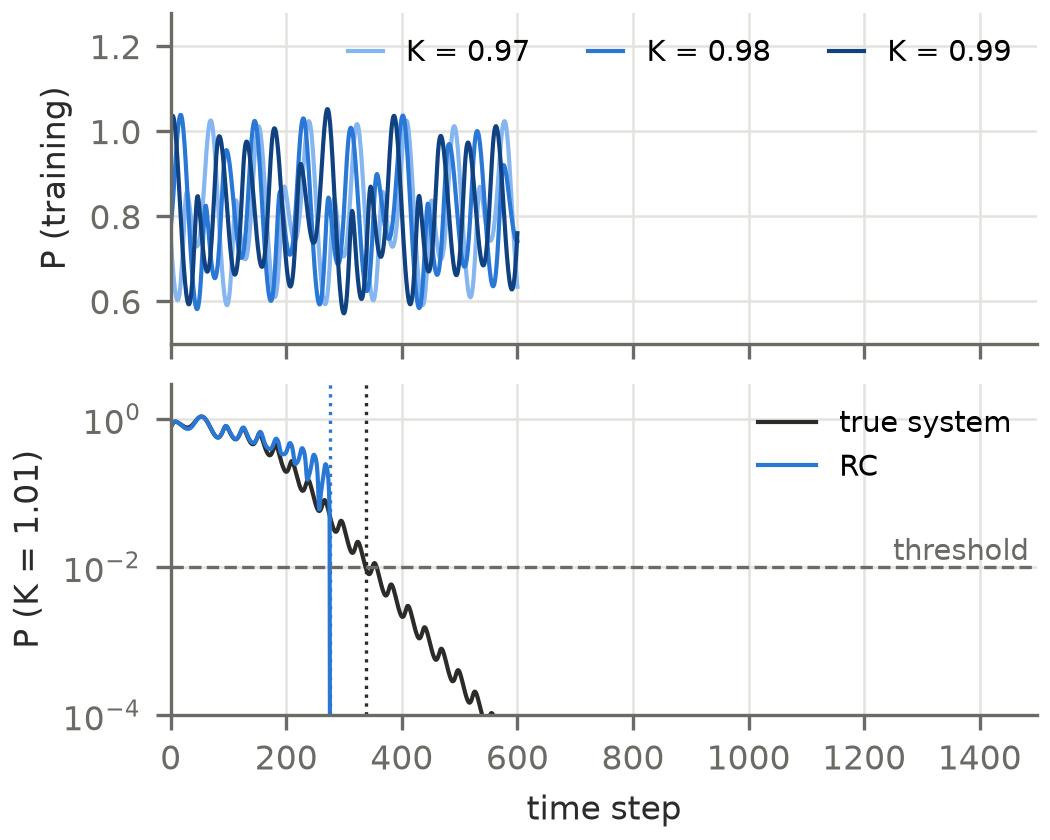}
\caption{Predator density}
\end{subfigure}\hfill
\begin{subfigure}[t]{0.49\textwidth}
\centering
\includegraphics[width=\linewidth]{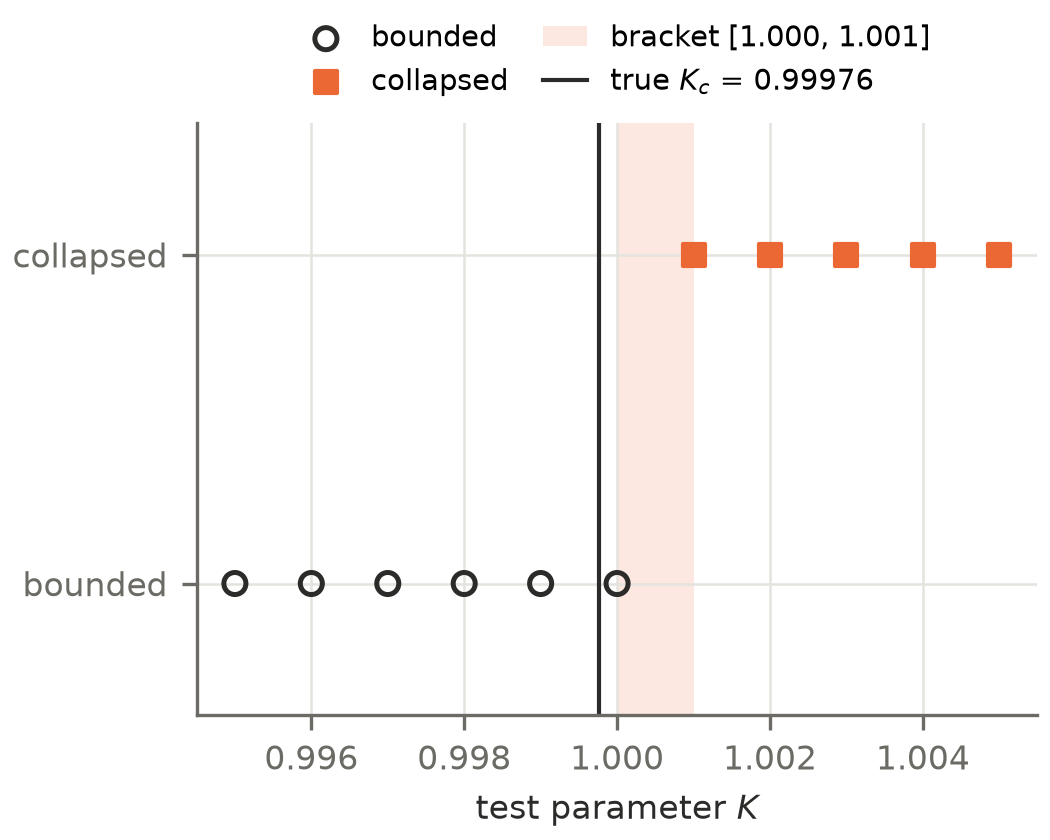}
\caption{Critical-point scan (horizon 1200)}
\end{subfigure}
\caption{Food chain. (a) Top: true predator density $P(t)$ at the three training values of
$K$. Bottom: $P(t)$ at $K = 1.01$ for the true system (black) and the RC prediction (blue)
from the same initial state, with the extinction threshold $P = 0.01$ (dashed); dotted lines
mark where each crosses it. (b) Critical-point scan: the shaded interval is the bracket
between the last bounded and the first collapsed value, and the solid line is the true
$K_c \approx 0.99976$~\cite{kong2021}.}
\label{fig:food-chain}
\end{figure}

\paragraph{Food chain:} A single reservoir (seed 3), with collapse defined as predator
extinction ($P < 0.01$), stays bounded at $K = 0.99$. At $K = 1.01$ it predicts extinction
after 275 steps, against 337 steps for the true system started from the same state
(Fig.~\ref{fig:food-chain}a). Its scan over $K \in [0.995, 1.005]$ with a horizon of 1200
steps is monotonic and brackets the transition between 1.000 and 1.001, just above the true
$K_c \approx 0.99976$ (Fig.~\ref{fig:food-chain}b). This result has not been repeated over
an ensemble of reservoirs.

\begin{figure}[htbp]
\centering
\includegraphics[width=\textwidth]{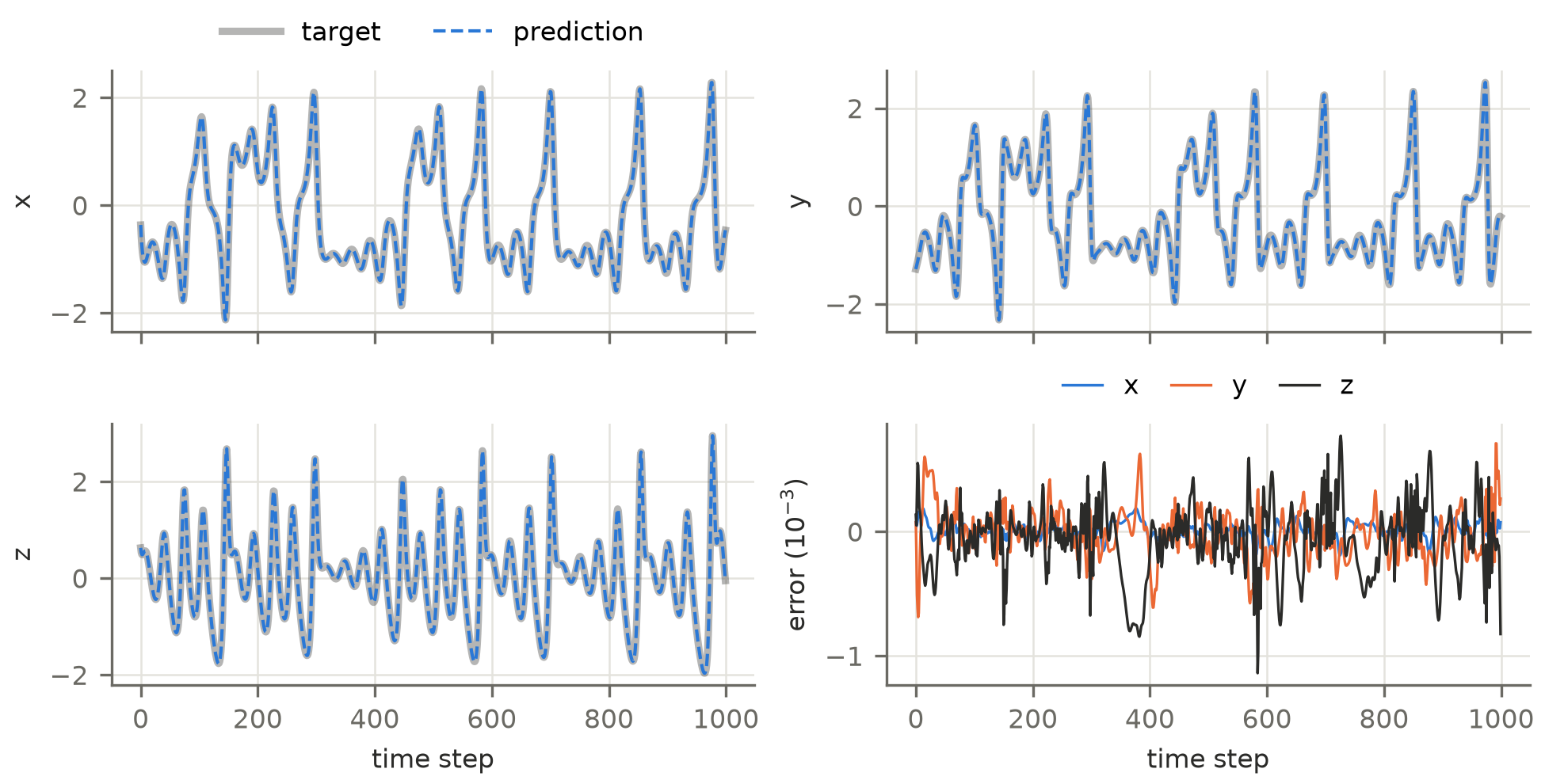}
\caption{One-step-ahead prediction of the Lorenz-63 system by a standard ESN with 500
units, on held-out data. The first three panels show the target and the prediction for the
standardized variables $x$, $y$, and $z$. The fourth panel shows the error, defined as the
target minus the prediction. The test NRMSE (Normalized Root Mean Square Error) is $2.15 \times 10^{-4}$.}
\label{fig:lorenz}
\end{figure}

\paragraph{Lorenz-63:} A standard ESN trained on the Lorenz-63
attractor~\cite{lorenz1963} closely tracks the held-out trajectory in one-step-ahead
prediction (Fig.~\ref{fig:lorenz}), validating the core reservoir implementation. This
example involves neither the parameter channel nor closed-loop rollout; it is therefore not
a test of parameter-aware prediction, and one-step-ahead accuracy alone does not establish
the long-term fidelity of the autonomous dynamics.

\section*{Reproducibility}

All results were produced with version 0.1.3 by the scripts below, in the repository's
\texttt{examples} directory. Each script fixes every random seed.

\paragraph{Scripts.}
\begin{itemize}[nosep]
    \item \texttt{ikeda\_blind\_tuning.py}: the blind hyperparameter search
    (Table~\ref{tab:tuning}); writes the selected configuration and the ten best candidates.
    \item \texttt{ikeda\_validation.py}: Fig.~\ref{fig:ikeda-validation} and all Ikeda
    numbers in the text, and the raw collapse steps of every run. An option switches from the
    blind configuration to the one tuned with post-critical information.
    \item \texttt{ikeda\_prediction.py}, \texttt{food\_chain\_prediction.py}: the
    single-model examples.
    \item \texttt{food\_chain\_figure.py}: Fig.~\ref{fig:food-chain}.
    \item \texttt{lorenz\_prediction.py}: Fig.~\ref{fig:lorenz}.
    \item \texttt{gpu\_benchmark.py}: Table~\ref{tab:gpu-bench} and the CPU/GPU agreement
    figures.
\end{itemize}

\paragraph{Initial conditions.}
\begin{itemize}[nosep]
    \item \textit{Training series:} simulated from the seeds in
    Table~\ref{tab:example-settings} after a burn-in of 1000 steps (Ikeda map) or 600 time
    units (food chain).
    \item \textit{Ikeda validation, true map:} 400 initial conditions drawn uniformly from
    $[0,1]^2$ (seed 0) and iterated 2000 steps at $\mu = 0.97$, so that they lie on the
    pre-critical attractor.
    \item \textit{Ikeda validation, RC:} rollouts over reservoirs are warmed up with the last
    50 training steps at $\mu = 0.97$; rollouts over initial conditions with 24 disjoint
    50-step windows, spaced 500 steps apart, of a separate true trajectory at $\mu = 0.97$
    (seed 7).
    \item \textit{Food chain:} the true post-critical trajectory starts from the same state as
    the RC rollout, the last training state at $K = 0.99$.
\end{itemize}

\section*{Quality Control}

The package ships a suite of 41 unit tests, run with pytest, covering fitting, prediction,
collapse classification, critical-point scans with their brackets, and censored
ensembles of the parameter-aware model; the simulators and training sets of all three
benchmark systems; the standard ESN, reservoir, ridge readout, and error metrics;
reproducibility of reservoirs built from the same seed; and the hyperparameter search.
Four of the tests concern GPU acceleration. Two check that CPU and GPU results agree and run
only on a machine with a CUDA device. The other two check that requesting the GPU without
one raises an error and run only on a machine without a CUDA device. On any machine,
therefore, 39 tests run and 2 are skipped.

\FloatBarrier
\section*{Conclusion}

\pkg{} packages the parameter-aware reservoir-computing method of
Kong et al.~\cite{kong2021} into a documented, tested, and installable Python library. Its
core is the PARC workflow: a single reservoir with a parameter channel is trained on time
series pooled across several parameter values and then run as an autonomous closed-loop
system at prescribed, previously unseen parameter values. The package also includes a
general-purpose echo state network, Ikeda-map, food-chain, and Kuramoto--Sivashinsky
benchmarks, hyperparameter search, and optional GPU acceleration for large reservoirs. It
is available on PyPI and on GitHub at
\url{https://github.com/jinchen7-cmd/Reservoir-Computing} under the Apache License 2.0.

The scientific application featured in this paper is the prediction of critical
transitions and system collapse from pre-critical data. For this task, the package
provides the tools needed to check its predictions: collapse classification that
separates physical collapse from numerical divergence, critical-point scans that report
their horizon and bracket, and ensembles with censored transient lifetimes. In a blind test
on the Ikeda map, with hyperparameters and collapse bound chosen from pre-critical data
only, the model correctly separates bounded from collapsing regimes near the transition,
but the ensemble places it about 0.005 too high in $\mu$, a substantial fraction of runs
does not collapse within the horizon, and transient lifetimes are reproduced only in order
of magnitude; tuning with post-critical information narrows the critical-point error but not
the lifetime discrepancy. For the food chain, a single reservoir brackets the transition
just above the true $K_c$, a result not yet repeated over an ensemble of reservoirs.

These results also delineate the practical limitations of the approach. Hyperparameter
selection is a dominant source of uncertainty, and single reservoirs are considerably less
reliable than ensembles, so predictions should be reported with their spread across
independently initialized reservoirs. Finite prediction horizons bias critical-point
estimates and censor long transients, and the state reached after a predicted collapse is
not a prediction of the physical post-collapse state. The present implementation requires
pre-critical training data at three or more parameter values, and the examples in this
paper address collapse prediction only; other applications of PARC, such as the
reconstruction of bifurcation diagrams or the anticipation of synchronization, are
established in the cited literature but are neither implemented as dedicated utilities nor
demonstrated here. Within these limits, \pkg{} provides a reproducible computational
foundation on which such studies of parameter-dependent nonlinear dynamics can be built.

\makeatletter
\immediate\write\@auxout{\string\citation{REVTEX42Control}}
\immediate\write\@auxout{\string\citation{apsrev42Control}}
\makeatother
\bibliographystyle{apsrev4-2}
\bibliography{references}

\end{document}